\def\year{2020}\relax
\documentclass[letterpaper]{article} 
\usepackage{aaai20}  
\usepackage{times}  
\usepackage{helvet} 
\usepackage{courier}  
\usepackage[hyphens]{url}  
\usepackage{graphicx} 
\usepackage{graphicx}  
\usepackage{booktabs}
\usepackage{subfigure}
\usepackage{amsmath,bm}
\usepackage{multirow}
\usepackage{amsfonts,amssymb}

\title{Are Noisy Sentences Useless for Distant Supervised Relation Extraction?}
\author{Yu-Ming Shang\textsuperscript{\rm 1,2,4}, He-Yan Huang\textsuperscript{\rm 1}, Xian-Ling Mao\textsuperscript{\rm 1}, Xin Sun\textsuperscript{\rm 1}, Wei Wei\textsuperscript{\rm 3} \\\\ 
\textsuperscript{\rm 1} School of Computer Science, Beijing Institute of Technology, Beijing, China\\ 
\textsuperscript{\rm 2} CETC Big Data Research Institute Co., Ltd., Guiyang, China,550022 \\
\textsuperscript{\rm 3} Huazhong University of Science and Technology, Hu'bei, China\\
\textsuperscript{\rm 4} Big Data Application on lmproving Government Governance Capabilities\\ National Engineering Laboratory Guiyang, China, 550022 \\
\{ymshang, hhy63, maoxl, sunxin\}@bit.edu.cn, Weiw@hust.edu.cn 
}
 
\begin{document}

\maketitle

\newcommand{\citet}[1]{\citeauthor{#1} \shortcite{#1}}
\newcommand{\citep}{\cite}
\newcommand{\citelp}[1]{\citeauthor{#1} \citeyear{#1}}

\begin{abstract}
	
	The noisy labeling problem has been one of the major obstacles for distant supervised relation extraction.
	Existing approaches usually consider that the noisy sentences are useless and will harm the model's performance. 
	Therefore, they mainly alleviate this problem by reducing the influence of noisy sentences, such as applying bag-level selective attention or removing noisy sentences from sentence-bags.
	However,  the underlying cause of the noisy labeling problem is not the lack of useful information, but the missing relation labels.
	Intuitively, if we can allocate credible labels for noisy sentences, they will be transformed into useful training data and benefit the model's performance.
	Thus, in this paper, we propose a novel method for distant supervised relation extraction, which employs unsupervised deep clustering to generate reliable labels for noisy sentences. 
	Specifically, our model contains three modules: a sentence encoder, a noise detector and a label generator. The sentence encoder is used to obtain feature representations. The noise detector detects noisy sentences from sentence-bags, and the label generator produces high-confidence relation labels for noisy sentences.
	Extensive experimental results demonstrate that our model outperforms the state-of-the-art baselines on a popular benchmark dataset, and can indeed alleviate the noisy labeling problem.

\end{abstract}

\section{Introduction}
	Relation Extraction, defined as the task of extracting structured relations from unstructured text, is a crucial task in natural language processing (NLP). 
	One of the main challenges of relation extraction is the lack of large-scale manually labeled data. Thus, \citet{mintz2009distant} proposed distant supervision to automatically construct training data. The assumption of distant supervision is that if two entities ($e_1, e_2$) have a relationship $r$ in knowledge graph, then any sentence that mentions the two entities might express the relation $r$.
	
	\begin{table}[t]
		\scalebox{0.67}{
			\begin{tabular}{@{}c|l|l|c|c@{}}
				\toprule
				& \multicolumn{1}{c|}{Sentence}                                                                                                                                            & Bag Label                     & Noise? & Correct Label \\ \midrule
				\multirow{7}{*}{Bag} & \begin{tabular}[c]{@{}l@{}}\#1:\textbf{ Barack Obama} was born in the \\ \textbf{United States}.\end{tabular}                                                         & \multirow{7}{*}{president of} & Yes    & born in       \\ \cmidrule(lr){2-2} \cmidrule(l){4-5} 
				& \begin{tabular}[c]{@{}l@{}}\#2: \textbf{Barack Obama} was the first \\ African American to be elected to the \\ president of the \textbf{United States}.\end{tabular} &                               & No     & president of  \\ \cmidrule(lr){2-2} \cmidrule(l){4-5} 
				& \begin{tabular}[c]{@{}l@{}}\#3: \textbf{Barack Obama} served as the 44th \\ president of the \textbf{United States} from \\ 2009 to 2017.\end{tabular}                &                               & No     & president of  \\ \bottomrule
			\end{tabular}
		}
		\caption{An example of sentence-bag annotated by distant supervision. ``Yes" and ``No" indicate whether or not each sentence is a noisy sentence. ``Correct Label" means the true relationship between the entity pair expressed in each sentence.}
		\label{noisy example}
	\end{table}
	
	Obviously, this assumption is too strong and will cause the noisy labeling problem.
	Since it only focuses on the existence of entities in text and knowledge graph, but cannot identify the one-to-one mapping between sentences and relations. 
	For example, as shown in Table~\ref{noisy example}, (\textit{Barack Obama, president of, United States}) is a relational triple in knowledge graph. Distant supervision will regard all sentences that contain [\textit{Barack Obama}]$_{e_1}$ and [\textit{United States}]$_{e_2}$ as the instance of relation ``\textit{president of}". 
	As a consequence, the first sentence which expresses the relation ``\textit{born in}" is wrongly labeled with relation ``\textit{president of}", and becomes a noisy sentence in the sentence-bag.
	
	Previous studies usually adopt Multi-Instance Learning (MIL) framework to address this problem~\cite{Riedel2010Modeling}. In this framework, the training and test process is proceeded at the sentence-bag level, where the sentence-bag contains all the sentences that mention the same triple ($e_1, r, e_2$).
	Existing MIL studies broadly fall into two categories: One is the \textbf{soft} decision methods, which tend to place soft weights on sentences to reduce the impact of noisy sentences~\cite{lin2016neural,yuan2019distant,yuan2019cross,ye-ling-2019-distant}. 
	The other is the \textbf{hard} decision methods that try to remove noisy sentences from sentence-bags to eliminate their influence~\cite{zeng2015distant,feng2018reinforcement,qin2018robust}. 
	
	However, previous de-noising methods ignore the essential cause of the noisy labeling problem --- the lack of correct relation labels. 
	To fill this gap, we try to solve this problem from the perspective of noisy sentences utilization, i.e., correcting their wrong labels.
	As shown in Table~\ref{noisy example},
	``\textit{Barack Obama was born in the United States}" is a noisy sentence in the sentence-bag. While it indeed expresses the relation ``\textit{born in}" between [\textit{Barack Obama}]$_{e_1}$ and [\textit{United States}]$_{e_2}$.
	Intuitively, if we can change its relation label from ``\textit{president of}" to ``\textit{born in}", it will be transformed into a useful training instance. This idea brings two advantages: (1) The negative influence of noisy sentences is reduced. (2) The number of useful training data is increased.

	In this paper, we propose a novel \textbf{D}eep \textbf{C}lustering based \textbf{R}elation \textbf{E}xtraction model, named DCRE, which employs unsupervised deep clustering to generate high-confidence labels for noisy sentences.
	More specifically, DCRE consists of three modules: a sentence encoder, a noise detector and a label generator.
	The sentence encoder is adopted to derive the sentence representation and shared by the other two modules. 
	The noise detector selects noisy sentences from sentence-bags according to the matching degree between sentences and the bag-level target relations. The sentence who scored below a certain threshold will be treated as noisy sentence.
	The label generator produces reliable labels for noisy sentences with the help of the deep clustering neural network.
	Because the results of unsupervised clustering may have errors, we further utilize clustering confidences as weights to scale the loss function.
	Experimental results show that our model performs better than the state-of-the-art baselines. Our contributions are as follows:
	\begin{itemize}
		\item Different from existing bag-level de-noising methods, our model tries to convert the noisy sentences as useful training data and can simultaneously decrease noisy data and increase useful data.
		\item To the best of our knowledge, it is the first work to apply unsupervised deep clustering to obtain more appropriate relation labels for noisy sentences.
		\item Extensive experiments show that our model outperforms the state-of-the-art baselines, and can effectively alleviate the noisy labeling problem.
	\end{itemize}

\section{Related Work}

	This paper proposes an unsupervised deep clustering based distant supervised relation extraction model. Related works to this paper mainly includes:
	
	\subsection{Distant Supervised Relation Extraction}
	
	Distant supervision~\cite{mintz2009distant} is proposed to obtain large-scale training data automatically and has become the standard method for relation extraction.
	However, the training data generated by distant supervision often contain amounts of noisy sentences. Therefore, noise-reduction has become the mainstream of distant supervised relation extraction. According to the way of processing noisy sentences, existing de-noising methods can be divided into three categories:
	
	The first category of methods tend to assign soft weights on sentences or sentence-bags. By conducting selective attention, it allows the model to focus more on the sentences of a higher quality and reduce the impact of noisy sentences. 
	For example, \citet{lin2016neural} employ attention mechanism by distributing different weights to each sentence to capture the bag representation. \citet{yuan2019distant} use non-independent and identically distributed relevance of sentences to obtain the weights of each sentence. \citet{yuan2019cross} utilize cross-relation cross-bag selective attention to reduce the impact of noisy sentences. \citet{ye-ling-2019-distant} consider both intra-bag and inter-bag attention in order to deal with noisy sentences at sentence-level and bag-level. 
	
	The second category of methods try to remove noisy sentences from sentence-bags through hard decision. For example, \citet{zeng2015distant} select the most correct sentence from each bag and ignore other sentences. \citet{feng2018reinforcement} employ reinforcement learning to train an instance selector and remove the wrong samples from sentence-bags. \citet{qin2018robust} also use reinforcement learning to process noisy sentences. Different from \cite{feng2018reinforcement}, they re-distribute noisy sentences into negative samples. 
	
	Different from the first two categories, the third type of approaches do not directly process noisy sentences during training stage. For example, \citet{takamatsu2012reducing} use syntactic patterns to identify the latent noisy sentences and remove them during the pre-processing stage. \citet{wang2018label} avoid using noisy relation labels and employs $e_2 - e_1$ as soft label to train the model. \citet{wu2019improving} propose a linear layer to obtain the connection between true labels and noisy labels. Then, conduct final prediction based on only the true labels.  
	
	Similar with the first two categories, the model proposed in this paper also directly processes noisy sentences during training stage. The main difference between DCRE and other methods is: DCRE tries to convert the noisy sentences into meaningful training data. It can simultaneously reduce the number of noisy sentences and increase the number of useful sentences. 

\subsection{Unsupervised Deep Clustering}

	There are broadly two types of deep clustering algorithms: 
	
	The first category of algorithms directly take advantage of the low dimensional features learned by other neural networks, and then run conventional clustering algorithm like \textit{k}-means. For example, \citet{tian2014learning} utilize auto-encoder to learn feature representations, and then obtain the clustering results by conducting \textit{k}-means algorithm.
	
	The second category tries to learn the feature representation and clustering in an end-to-end mechanism. Among these methods, Deep Embedded Clustering (DEC) is a specialized clustering technique~\cite{xie2016unsupervised}. This method employs a stacked auto-encoder learning approach. After obtaining the hidden representation of the auto-encoder by pre-training, the encoder pathway is fine-tuned by a defined Kullback-Leibler divergence clustering loss. \citet{guo2017improved} consider that the defined clustering loss will corrupt the feature space and the pre-train process is too complicated, so they keep the decoder remained and add a re-construction loss.
	Since then, there have been increasing algorithms based on such deep clustering framework~\cite{ghasedi2017deep,guo2017deep}.
	
	The deep clustering neural architecture proposed in our work falls into the second category. It utilizes the features produced by the pre-trained sentence encoder, and then, jointly optimizes the feature representation and clustering in an end-to-end way.

\section{Method}

\begin{figure}[t]
		\centering
		\includegraphics[width=1.01\columnwidth]{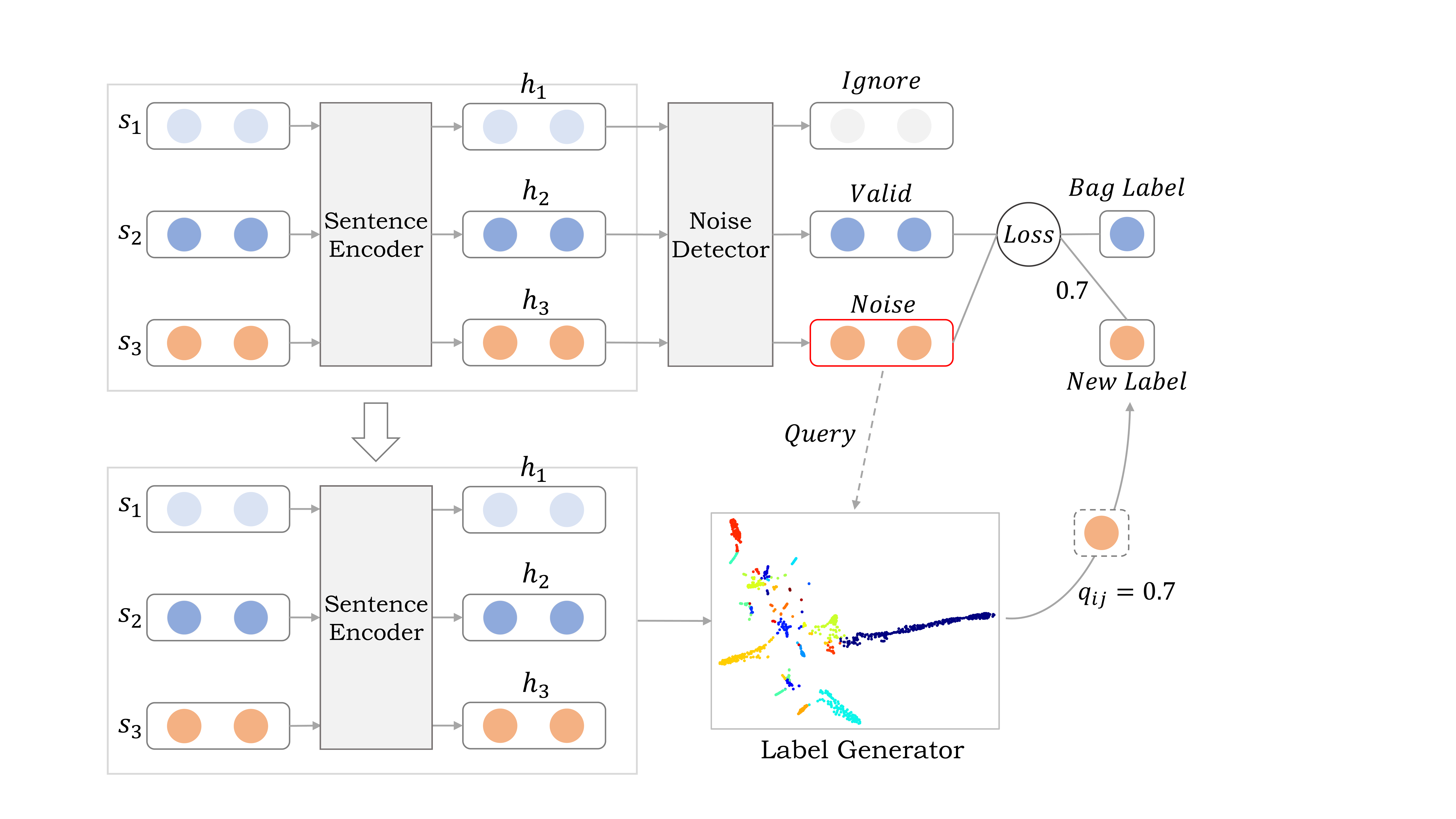} 
		\caption{The architecture of DCRE, illustrating the procedure of handling one sentence-bag which contains three sentences. }
		\label{model}
	\end{figure}

	In this section, we present our method of distant supervised relation extraction. The architecture of the neural network is illustrated in Figure~\ref{model}. It shows the procedure of handling one sentence-bag. Our model contains three main modules: a sentence encoder, a noise detector and a label generator.  In the following, We first give the task definition and notations. Then, we provide detail formalization of the three modules.

\subsection{Task Definition and Notation}
	We define the relation classes as $\mathbb{R} = \{r_1, r_2, ..., r_k\}$, where $k$ is the number of relations. Given a bag of sentences $\mathbb{S}_b = \{s_1, s_2, ..., s_b\}$ consisting of $b$ sentences and an entity pair ($e_1, e_2$) presenting in all sentences. The purpose of distant supervised relation extraction is to predict the relation $r_i$ of sentence-bag $\mathbb{S}_b$ according to the entity pair ($e_1, e_2$). Therefore, the relation extraction is defined as a classification problem.

\subsection{Sentence Encoder}
	When conducting relation extraction, sentences need to be transformed into low-dimensional vectors. We first transform a sentence into a matrix with word embeddings and position embeddings. Then, a Piece-wise Convolution (PCNN)~\cite{zeng2015distant} layer is used to obtain the final sentence representation.
	
\subsubsection{Word Representation}

	In a sentence $s$, each word $w_i$ is first mapped into a $d_w$-dimensional word embedding $\bm{v}_i$. The position features (PFs) proposed by~\cite{zeng2014relation} are adopted in this work to specify the target entity pair and make the model pay more attention to the words close to the target entities. PFs are a series of relative distances from the current word to the two entities. The position embedding $\bm{p}_{i}^{e_1}, \bm{p}_{i}^{e_2}$ are low dimensional vectors of PFs.
	The final representation $\bm{x}_i$ of each word $w_i $ is the concatenation of the word embedding and two position embeddings [$\bm{v}_i; \bm{p}_{i}^{e_1}; \bm{p}_{i}^{e_2}$]. Then the input sentence representation is:
	
	\begin{large}
		\begin{equation}
		\mathrm{\textbf{X}}  = \bm{x}_1, \bm{x}_2, ..., \bm{x}_{n_l},
		\end{equation}
	\end{large}
	where $n_l$ is the length of the sentence.

\subsubsection{PCNN}

	We employ PCNN as our feature extractor, which mainly consists of two parts: one-dimensional convolution and piece-wise max-pooling.
	
	One-dimensional convolution is an operation between a matrix of weights $\mathrm{\textbf{W}} $, and a matrix of inputs viewed as a sequence $\mathrm{\textbf{X}} $. $\mathrm{\textbf{W}} $ is regarded as the \textit{filter} for the convolution and $\bm{x}_i$ is a input vector associated with the $i$-th word in the sentence. In general, let $\bm{x}_{i:j}$ refer to the concatenation of $\bm{x}_i$ to $\bm{x}_j$, $w$ refer to filter size. The convolution is to take the dot product of the vector $\mathrm{\textbf{W}}$ with each $w$-gram in the sentence $\mathrm{\textbf{X}}$ to obtain another sequence $\bm{m}_i$:
	
	\begin{equation}
	\bm{m}_i = \mathrm{\textbf{W}}^T \bm{x}_{i-w+1:i}.
	\end{equation}
	
	The number of $\bm{m}_i$ is \textit{$n_l-w+1$}. In our model, each sentence is padded with padding-elements such that the number of vector $\bm{m}_i$ is equal to the length of the sentence $n_l$. The convolution result is a feature map matrix $\mathrm{\textbf{M}}$ = \{$\bm{m}_1, \bm{m}_2, ..., \bm{m}_{n_l}$\}. The number of feature map $\mathrm{\textbf{M}}_i$ is $n_f$, where $n_f$ is the number of filters.
	
	Piece-wise max-pooling is used to capture the structural information of sentences. After convolution layer, each feature map $\mathrm{\textbf{M}}_i$ is divided into three parts \{ $\mathrm{\textbf{M}}_{i1}$, $\mathrm{\textbf{M}}_{i2}$, $\mathrm{\textbf{M}}_{i3}$ \} by the position of two entities. Then, the max-pooling operation is performed on the three parts separately. The final sentence representation $\bm{h}$ is the concatenation of all vectors:
	\begin{equation}
	\bm{h} = [\bm{p}_{i1}, \bm{p}_{i2}, \bm{p}_{i3}],
	\end{equation}
	where $\bm{p}_{ik} = max$($\mathrm{\textbf{M}}_{ik}$), $k \in \{1, 2, 3\}$. To prevent over-fitting, the dropout strategy~\cite{srivastava2014dropout} is applied to sentence representation matrices.

\subsection{Noise Detector}

	The purpose of noise detector is to select noisy sentences from sentence-bags and feed them to the label generator.
	Let $\mathrm{\textbf{H}}_b = \{\bm{h}_1, \bm{h}_2, ... , \bm{h}_b\}, \mathrm{\textbf{H}}_b \in \mathrm{\textbf{R}}^{b \times d_s}$ represents the representation of a sentence-bag, $\bm{h}_i$ is the $d_s$-dimensional sentence representation produced by sentence encoder, and $b$ is the number of sentences. Let $\mathrm{\textbf{L}} = \{\bm{l}_1, \bm{l}_2, ..., \bm{l}_k\}, \mathrm{\textbf{L}} \in \mathrm{\textbf{R}} ^{k \times d_s}$ denotes the representations of all the relations. Firstly, a simple dot product between the sentence representation $\bm{h}_i$ and the vector of bag-level relation label $\bm{l}_j$ is adopted to calculate the coupling coefficient as:
	
	\begin{equation}
	a_i = \bm{h}_i  \bm{l}_j^T.
	\end{equation}
	
	Then, the coupling coefficient is normalized at the bag-level through a softmax function:
	
	\begin{equation}
		a_i = \frac{exp(a_i)}{\sum_{b} exp(a_i)},
	\end{equation}

	where each $a_i$ corresponds to the matching degree between each sentence and the target relation. 
	It represents the possibility that the original relation label is correct for the current sentence.
	We set a threshold $\phi$ to detect noisy sentences, and the sentence whose coupling coefficient is less than $\phi$ will be regarded as a noisy sample.	
	
	However, we can't guarantee that the sentences with a higher coefficient are not wrongly labeled. 
	For this indetermination, our solution is to use the currently deterministic sentences. In a sentence-bag, we consider the best scored sentence as valid sample. The sentences that are neither determined as noisy sentences (scored below $\phi$), nor determined as valid samples (best score) will be ignored. 
	The reasons behind this operation are: (1) If the best scored sentence indeed expresses the target relation, it is consistent with the \textit{expressed-at-least-once} assumption~\cite{Riedel2010Modeling}. This assumption believes that in a sentence-bag, \textit{at least one sentence} might express the target relation.  (2) If the best scored sentence is a noisy sample, in other words, all the sentences in the bag are noisy samples and the sentence-bag is a \textit{noisy bag}~\cite{ye-ling-2019-distant}, ignoring uncertain samples is actually removing the noisy sentences. In both cases, re-labeling high-confidence noisy sentences will benefit the model's performance.

\subsection{Label Generator}

	The label generator provides high-confidence relation labels for noisy sentences based on the deep clustering neural network. Let $\mathrm{\textbf{H}} = \{\bm{h}_1, \bm{h}_2, ... , \bm{h}_n\}, \mathrm{\textbf{H}} \in \mathrm{\mathbf{R}}^{n \times d_s}$ denotes the representations of all sentences produced by the pre-trained sentence encoder, and $\mathrm{\textbf{L}} = \{\bm{l}_1, \bm{l}_2, ..., \bm{l}_k\}, \mathrm{\textbf{L}} \in \mathrm{\mathbf{R}} ^{k \times d_s}$ denotes the pre-trained relation matrix. Firstly, we project sentence representations into relation feature space:
	\begin{equation}
	\mathrm{\textbf{C}} = \mathrm{\textbf{H}} \mathrm{\textbf{L}}^T + \bm{b},
	\end{equation}
	
	where $\bm{b}$ is a bias. This operation can be viewed as an attention with all the relations as query vectors to calculate the relation-aware sentence representations.  Then, we feed $\mathrm{\textbf{C}}$ into the clustering layer, which maintains cluster centers \{$\bm{\mu}_i$\}$_{i=1}^{n_c}$ as trainable weights, where $n_c$ is the cluster number. We use the Student's $t$-distribution~\cite{maaten2008visualizing} as a kernel to measure the similarity $q_{ij}$ between the feature vector $\bm{c}_i$ and the cluster center $\bm{\mu}_j$:
	
	\begin{equation}
	q_{ij} = \frac{ (1 + \| \bm{c}_i - \bm{\mu}_j \|^2)^{-1}}{\sum_j (1 + \| \bm{c}_i - \bm{\mu}_j \|^2)^{-1}},
	\label{scale}
	\end{equation}
	
	where $q_{ij}$ is the similarity between the projected sentence vector $\bm{c}_i$ and the cluster center vector $\bm{\mu}_j$. It also can be interpreted as the probability of assigning the sentence $s_i$ with relation label $r_j$.  
	
\begin{figure}[t]
	\centering
	\includegraphics[width=0.95\columnwidth]{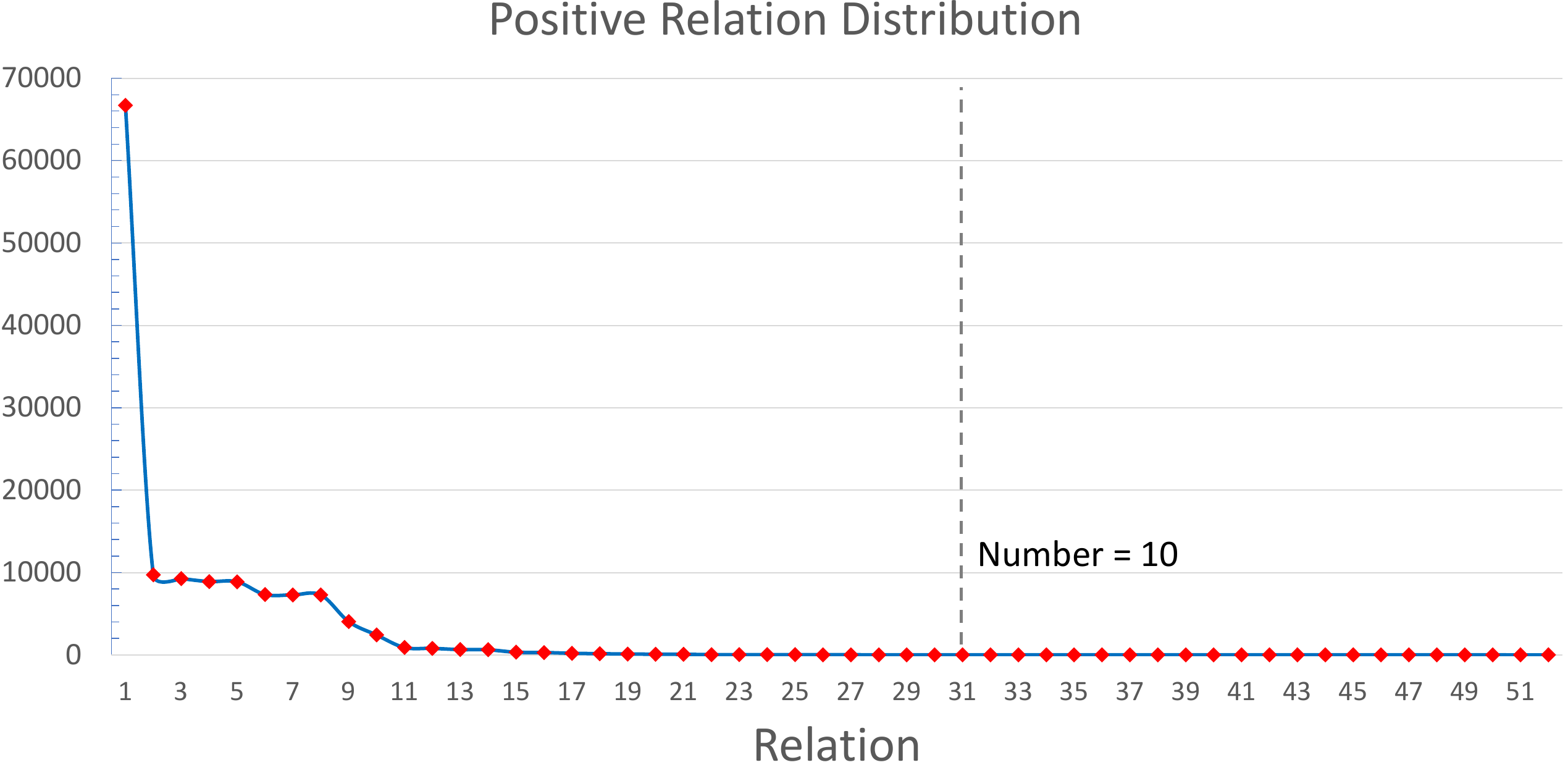} 
	\caption{The distribution of 52 positive relations (exclude \textit{NA}) in NYT-10 dataset. The horizontal axis shows different relations sorted by the number of occurrence. The vertical axis shows the number of sentences in training set. The vertical line indicates that the relation whose id is 31 appears 10 times in the training set.  }
	\label{generator}
\end{figure}

	The loss function of deep clustering is defined as a Kullback-Leibler divergence:
	
	\begin{equation}
	\mathcal{L} = KL(P \| Q) = \sum_i \sum_j p_{ij} log{\frac{p_{ij}}{q_{ij}}},
	\end{equation}
	
	where $P$ is the target distribution. The relations in NYT-10 follow a long-tail distribution, as shown in Figure~\ref{generator}. To alleviate this data imbalance problem, we use the same $P$ with~\cite{xie2016unsupervised}, defined as: 
	
	\begin{equation}
	p_{ij} = \frac{q_{ij}^2 / \sum_i q_{ij}}{\sum_j (q_{ij}^2 / \sum_i q_{ij})}.
	\end{equation}

	This target distribution can normalize the loss contribution of each centroid to prevent large clusters from distorting the hidden feature space.
	
	Noted that we only generate new labels for positive samples, i.e., the samples whose original labels are not \textit{NA} (No relations). Because the representations of sentences which express no relations are always diverse and it's difficult to find correct labels for them. Allowing negative samples to be re-labeled will produce more noisy sentences. On the contrary, a positive sample is re-labeled as \textit{NA} means that the noisy sentence is removed.

\subsection{Scaled Loss Function}

	Because there is not any explicit supervision for the noisy data, it's difficult to know whether the clustering result of each sentence is correct. Thus, the new labels produced by the label generator may still be wrong. 
	
	To tackle this problem, as mentioned above, we set a threshold $\phi$ and select sentences with high-confidence to be noisy samples. Further more, we introduce a scaling factor $q_{ij}$ as weight to scale the cross-entropy~\cite{shore1980axiomatic} loss function. The $q_{ij}$ is obtained by equation~(\ref{scale}), which denotes the probability of the $i$-th sentence is belonged to the $j$-th relation cluster. 
	This scaling factor makes the new labels have different influence on the model according to their clustering confidence.
	Finally, the object function is defined as:
\begin{small}
	\begin{equation}
	\begin{split}
	\mathcal{J}\left( \theta \right) = -\sum_{(x_i,y_i) \in \mathbb{V}} log { p(y_i|x_i ; \boldsymbol{\Theta})} \\ - \lambda\sum_{(x_i,y_i) \in \mathbb{N}} q_{ij} log { p(y_j|x_i ; \boldsymbol{\Theta})},
	\end{split}
	\end{equation}
\end{small}
	where $(x_i, y_i)$ is a training instance, means the target relation label of sentence $x_i$ is $y_i$. $y_j$ indicates that the new label for $x_i$ is $y_j$, and $y_j \neq y_i$. $\lambda$ is the coefficient that balances the two terms. $\mathbb{V}$ is the best scored samples, $\mathbb{N}$ is the noisy samples, $\boldsymbol{\Theta}$ indicates all parameters of the model. 

\section{Experiments}

	Our experiments are designed to demonstrate that DCRE can alleviate the noisy labeling problem. In this section, we first introduce the dataset and evaluation metrics. Second, we show the experiment setup. Third, we compare the performance of our model with several state-of-the-art approaches. Fourth, we make parameter analysis of the threshold $\phi$. Finally, we show some details of clustering results.

\begin{figure*}[t]
	\centering
	\includegraphics[width=1.8\columnwidth]{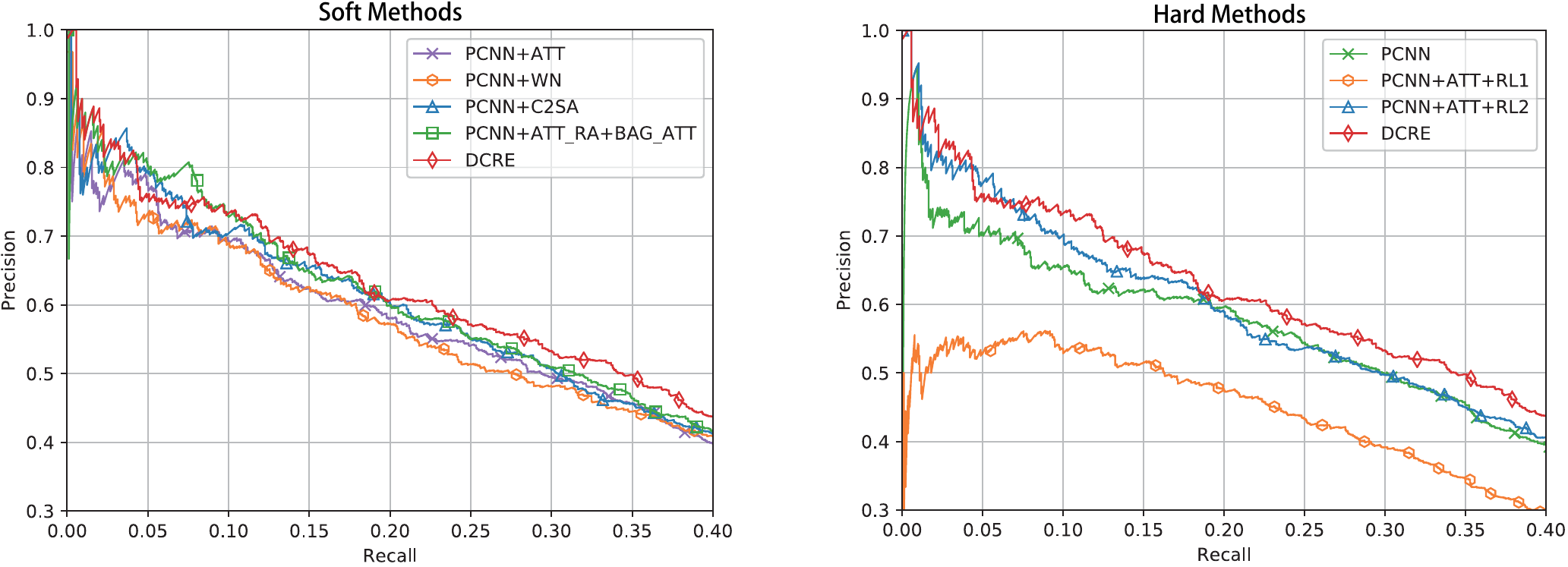}
	\caption{Comparison results with soft methods (left) and hard methods (right).}
	\label{experiment result}
\end{figure*}

\subsection{Data and Evaluation Metrics}

	We evaluate the proposed method on a widely used dataset NYT-10~\cite{Riedel2010Modeling} which was constructed by aligning relation facts in Freebase~\cite{bollacker2008freebase} with the New York Times (NYT) corpus. Sentences from 2005-2006 are used for training and sentences from 2007 are used for testing. Specifically, it contains 522,611 sentences, 281,270 entity pairs, and 18,252 relational facts in the training data; and 172,448 sentences, 96,678 entity pairs and 1,950 relational facts in the test data. There are 53 unique relations including a special relation \textit{NA} that signifies no relation between the entity pair.
	
	Following the previous works~\cite{yuan2019cross,yuan2019distant,ye-ling-2019-distant}, we evaluate our model and baselines in the held-out evaluation and present the results with precision-recall curves. 
	In held-out evaluation, the relations extracted from testing data are automatically compared with those in Freebase. It is an approximate measure of the model without requiring costly human evaluation. 

\subsection{Experiment Setup}
	During training, we first generate the new relation labels for noisy sentences by unsupervised deep clustering, and then, train the whole model.
	When conducting clustering, we employed \textit{k}-means to initialize the cluster centers for faster convergence. Both over-sampling and under-sampling strategies are applied to highlight the importance of positive samples and alleviate the data imbalance problem. For every positive sample, we obtain multiple clustering results and determine its final category by voting. Besides, we ignored 6 long-tail relations that appear less than 2 times. \footnote{The ignored relations are: \\ 
	{\small /business/shopping\_center\_owner/shopping\_centers\_owned},\\
	{\small /location/fr\_region/capital},\\  
	{\small /location/mx\_state/capital},\\
	{\small /business/shopping\_center/owner},\\
	{\small /location/country/languages\_spoken},\\
	{\small /base/locations/countries/states\_provinces\_within.}}
	
	For all the baselines, during training, we follow the settings used in their papers. Table~\ref{parameters setting} shows the main parameters used in our DCRE. 

\begin{table}[h]
	\centering
	\begin{tabular}{lrrrrr}  
		\toprule
		Setting  & Number \\
		\midrule
		Kernel size &   3\\
		Feature maps & 230  \\
		Word embedding dimension & 50 \\
		Position embedding dimension& 5\\
		Pre-train learning rate & 0.4\\
		Clustering learning rate & 0.004\\
		Model learning rate & 0.1\\
		Threshold $\phi$ & 0.1\\
		Dropout & 0.5\\
		Coefficient $\lambda$ & 0.6\\
		\bottomrule
	\end{tabular}
	\caption{Parameters Setting}
	\label{parameters setting}
\end{table}

\subsection{Experiment Results}
\subsubsection{Baselines}
	The model proposed in this work directly process noisy sentences during the training stage, so we select seven related works as baselines. Among these methods, PCNN+ATT, PCNN+WN, PCNN+C2SA and  PCNN+ATT\_RA+BAG\_ATT are soft decision methods, PCNN, PCNN+ATT+RL1 and PCNN+ATT+RL2 are hard decision methods.
\begin{enumerate}
	\item \textbf{PCNN+ATT}: \citet{lin2016neural} propose a selective attention over sentences based on PCNN sentence encoder.
	
	\item \textbf{PCNN+WN}: \citet{yuan2019distant} propose a non-independent and identically distributed relevance to capture the relevance of sentences in the bag.
	
	\item \textbf{PCNN+C2SA}: \citet{yuan2019cross} propose cross-relation cross-bag selective attention to handle noisy sentences. 
	
	\item \textbf{PCNN+ATT\_RA+BAG\_ATT}: \citet{ye-ling-2019-distant} propose intra-bag and inter-bag attention to deal with noisy sentences at both sentence-level and bag-level. It is the state-of-the-art method. 
	
	\item \textbf{PCNN}: \citet{zeng2015distant} propose a method to select the most well labeled sentence from sentence-bag, and ignore the other sentences.
	
	\item \textbf{PCNN+ATT+RL1}: \citet{feng2018reinforcement} propose a reinforcement method to remove noisy sentences from sentence-bags. 
	
	\item \textbf{PCNN+ATT+RL2}: \citet{qin2018robust} also propose a reinforcement model. Different from~\cite{feng2018reinforcement}, it redistributes noisy sentences into negative examples.
	
\end{enumerate}

	We implement PCNN+ATT, PCNN+WN, PCNN, and our DCRE. 
	For the sake of fairness, we evaluate PCNN+C2SA\footnote{https://github.com/yuanyu255/PCNN\_C2SA}, PCNN+ATT\_RA+BAG\_ATT\footnote{https://github.com/ZhixiuYe/Intra-Bag-and-Inter-Bag-Attentions}, PCNN+ATT+RL2\footnote{https://github.com/Panda0406/Reinforcement-Learning-Distant-Supervision-RE} with the codes provided by authors, and replace their training data with the one has 522,611 training sentences. For PCNN+ATT+RL1, we use the source code provided by Open-NRE\footnote{ https://github.com/thunlp/OpenNRE }.

\subsubsection{Overall Performance of DCRE}

	The overall performance of DCRE compared with the seven baselines is shown in Figure~\ref{experiment result}. The left sub-figure shows the comparison results with soft decision methods. The right sub-figure exhibits the comparison results with hard decision methods. We can find that, our DCRE captures the best performance among all the baselines. 
	
	Compare DCRE with four soft decision methods, as shown in the left part of Figure~\ref{experiment result}, we have the following observations: (1) DCRE performs much better than PCNN+ATT and PCNN+WN. This illustrates that assigning low weights to noisy sentences can only reduce its negative influence, but cannot eliminate the impact of noisy sentences. 
	(2) DCRE performs slightly better than PCNN+ATT\_RA+BAG\_ATT and PCNN+C2SA. 
	Different from PCNN+ATT, PCNN+WN and DCRE, these two methods obtain the final representation according to the \textit{super-bag}~\cite{yuan2019cross}, so that they utilize a wider range of information.
	While DCRE is still better than them. This demonstrates that the motivation of our work that finding high-confidence labels for noisy sentences is effective.
	
	The right part of Figure~\ref{experiment result} shows the results of comparing DCRE with three hard decision models. Ideally, the instance selector trained by reinforcement learning can remove all the noisy sentences and the hard decision methods should perform better than the soft decision methods. However, it can be observed that there is an obvious margin between DCRE and the other methods. We believe that this is mainly because: (1) As the principle of held-out evaluation, there are also noisy sentences in the test data. Therefore, it's difficult to use the evaluating results as reward to train the instance selector. (2) Deleting noisy sentences can indeed eliminate their influence, but ignores the useful information contained in noisy sentences. This results have further verified our intuition for re-labeling the noisy sentences and convert them into useful training data.

\begin{figure}[!t]
	\centering
	\includegraphics[width=0.85\columnwidth]{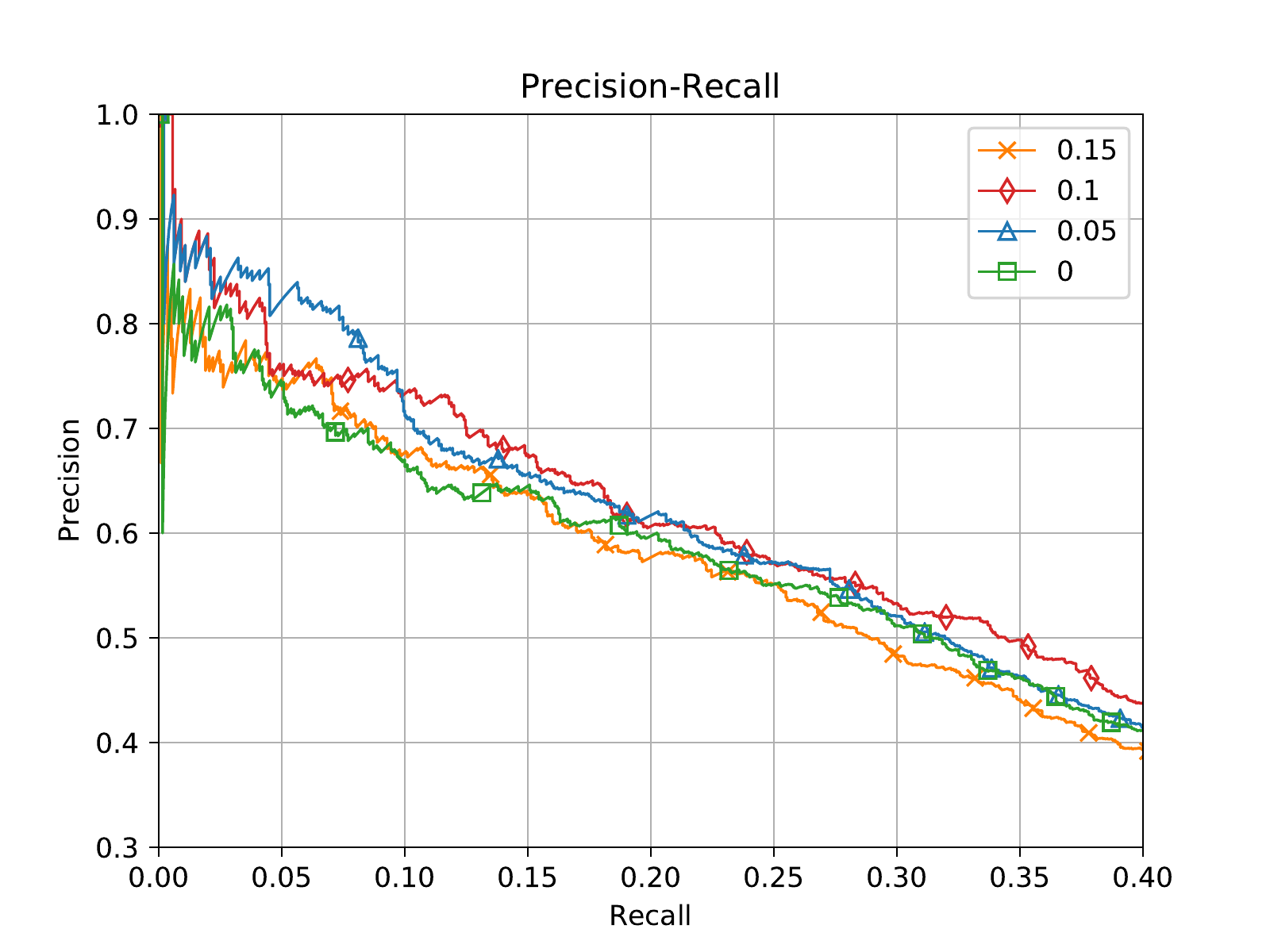}
	\caption{Effect of the threshold $\phi$.}
	\label{threshold}
\end{figure}

\begin{figure*}[!t]
	\centering
	\includegraphics[width=2\columnwidth]{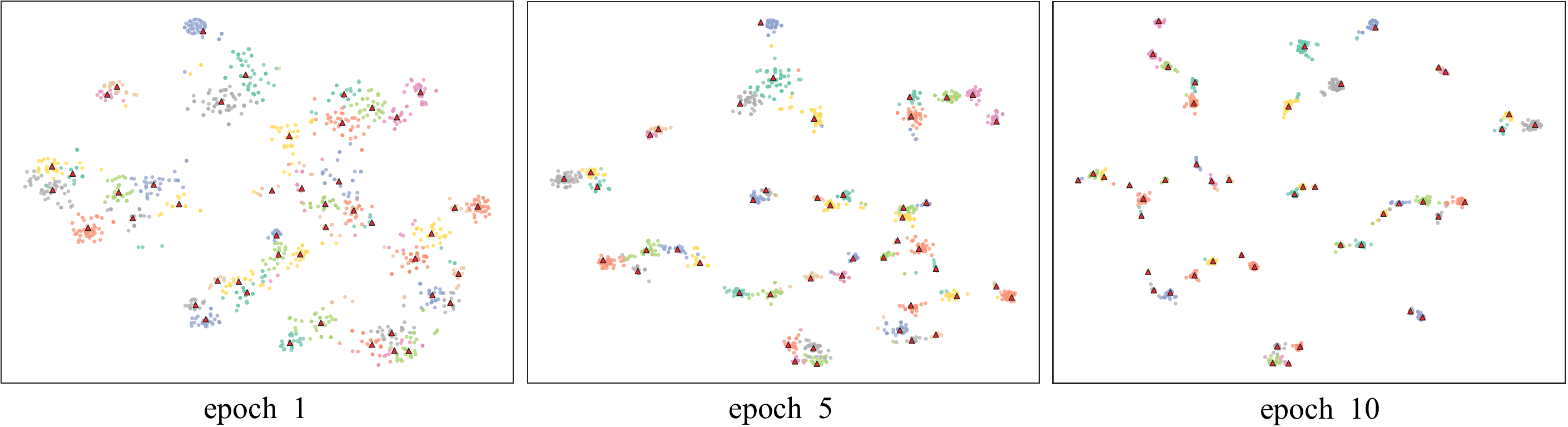}
	\caption{$t-$SNE visualization of clustering results on subset of NYT-10. The red triangles are cluster centers.}
	\label{clusteirng}
\end{figure*} 

\begin{table*}[t]
	\scalebox{0.72}{
		\begin{tabular}{@{}|c|c|l|l|l|c|@{}}
			\toprule
			ID & Entity pair               & \multicolumn{1}{c|}{Sentence}                                                                                                                                                                                                                                                                                                                                                                & \multicolumn{1}{c|}{Original label                          }                                           & \multicolumn{1}{c|}{Generated label }                                                          & Correct? \\ \midrule
			1  & (China,Beijing)           & \begin{tabular}[c]{@{}l@{}}\textbf{Beijing} has tried to enlist the support of Uzbekistan in fighting \\ Islamic separatism in \textbf{China}'s western region of Xinjiang, while \\ also lining up secure supplies of oil and gas.\end{tabular}                                                                                                                                         & /location/location/contains                                                      & \begin{tabular}[c]{@{}l@{}}/location/cn province \\ /capital\end{tabular} & No       \\ \midrule
			2  & (Italy, Rome)             & \begin{tabular}[c]{@{}l@{}}Mr. Tomassetti's companies are named after L'Aquila, \textbf{Italy}, his \\ birthplace 58 miles northeast of \textbf{Rome}.\end{tabular}                                                                                                                                                                                                                     & /location/country/capital                                                        & \underline{/location/location/contains }                                            & Yes      \\ \midrule
			3  & (Saddam Hussein, Iraq)    & \begin{tabular}[c]{@{}l@{}}As national journal reported in April, it was Senator Roberts \\ who stated as the \textbf{Iraq} war began that the U.S. had ``human \\ intelligence that indicated the location of \textbf{Saddam Hussein}."\end{tabular}                                                                                                                                 & \begin{tabular}[c]{@{}l@{}}/people/deceased\\ person/place of death\end{tabular} &\underline{ /people/person/place lived              }                                & Yes      \\ \midrule
			4  & (Edith Sitwell, England) & \begin{tabular}[c]{@{}l@{}}His first book was published privately in his own country and \\then by a major publisher in \textbf{England}, where he had many\\ supporters in the literary world, most notably \textbf{Edith Sitwell} and\\ Angus Wilson.\end{tabular} & \begin{tabular}[c]{@{}l@{}} /people/person/nationality\end{tabular}          & \begin{tabular}[c]{@{}l@{}}/people/person/place of birth\end{tabular} & No       \\ \midrule
			5  & (Louisiana, New Orleans)  & \begin{tabular}[c]{@{}l@{}}The book, by a \textbf{New Orleans} resident, John M. Barry, describes\\ the history and politics behind a flood that killed 1,000 people\\ and displaced 900,000 from \textbf{Louisiana} to Illinois.\end{tabular}                                                                                                                                          & /location/location/contains                                                      & \multicolumn{1}{c|}{\underline{NA}}                                                                                            & Yes      \\ \bottomrule
		\end{tabular}
	}
	\caption{Five sentences randomly selected in NYT-10 dataset. The text in \textbf{bold} represents the entity, the text \underline{underline} represents the relation label is correct.}
	\label{case}
\end{table*}

\subsubsection{Effect of the Threshold}
	The most important hyper-parameter of DCRE is the threshold $\phi$. To analyze how the $\phi$ affects the performance, we conduct experiments by selecting $\phi$ in the set \{0.15, 0.1, 0.05, 0\}, and the results of different thresholds are shown in Figure~\ref{threshold}. It can be found that the setting of $\phi$ is a process of reconciling contradictions and the model performs best when $\phi = 0.1$.
	The reasons behind this phenomenon are: 
	(1) A large $\phi$ means that some relatively high scored but not valid samples will be treated as noisy sentences. In other words, the original relation label and the new relation label of these sentences are all probably wrong. 
	(2) A relatively small $\phi$ indicates that the filtered sentences are more likely to be noisy sentences. While in this situation, the recall rate is too low.
	(3) When $\phi = 0$, the model is equal to PCNN which only utilizes the best scored sentence. 
	
\subsubsection{Clustering Result}
	We further verify the effectiveness of our deep clustering neural network by visualizing the clustering results during training. 
	We set the number of clusters as 47, excluding 6 long-tail relations which appear less than 2 times in training data. 
	We don't remove more long-tail relations because the noisy sentences labeled with other relations may be clustered into these clusters.  
	
	We randomly select 1000 sentences that covering all the 47 categories and visualize them by $t-$SNE~\cite{maaten2008visualizing}, as shown in Figure~\ref{clusteirng}. It can be found that there is a clear boundary between different clusters in epoch 1, but this clustering result cannot be used to re-label the noisy sentences. Because the distance between the points of the same cluster is too large so that the confidence of the new labels is relatively low. 
	It can be seen from the visualization, from left to right, the ``shape" of each cluster is becoming more and more compact and the clusters are becoming increasingly well separated. Accordingly, the confidence of the new label is becoming more high.
	
	In epoch 10, it can be found that some points of different clusters are close to each other. This phenomenon is consistent with reality, i.e., a sentence may express more than one relations. For example, \textit{[Barack Obama]$_{e_1}$ is the 44th president of the [United States]$_{e_2}$} is a high-quality sentence for relation ``\textit{president of}" and a low-quality sentence for relation ``\textit{live in}". Ideally, the sentence should be grouped into both clusters. In this paper, we only consider one high-confidence relation label for each noisy sentence.
	
	Furthermore, we randomly select five re-labeled sentences during training whose new relation labels are different from original labels to show the capabilities of noisy detector and label generator. Their target entity pairs, original labels and generated labels are illustrated in Table~\ref{case}. It can be found that: 
	(1) The original labels of the five sentences are wrong. This proves the validity of the threshold $\phi$.
	(2) The correct label for sentence 1 is \textit{/location/country/capital}. The original label is wrong because distant supervision cannot identify that the entity [\textit{Beijing}] represents the \textit{Chinese government}. The generated label is wrong mainly due to the word \textit{China}. 
	(3) Exactly speaking, the correct label for sentence 4 is \textit{people/person/place\_lived}. Its original label is \textit{/people/person/nationality} and generated label is \textit{/people/person/place\_of\_birth}. The three relations have inner connections so that it's difficult for the model to find the correct one. 
	
	\section{Conclusion and Future Work}
	In this paper, we proposed an unsupervised deep clustering based distant supervised relation extraction model. Different from conventional methods which focus on reducing the influence of noisy sentences, our model tries to find new relation labels for noisy sentences and convert them into useful training data. Extensive experimental results show that the proposed method performs better than comparable approaches, and can indeed alleviate the noisy labeling problem in distant supervised relation extraction.
	In the future, we will explore the following directions:

\begin{itemize}
	
	\item Our clustering algorithm assigns one relation label for each noisy sentence. While in reality, one sentence may express multiple relations. We will consider multi-class clustering in the future.
	
	\item  The threshold $\phi$ is very important in DCRE. We will next develop an end-to-end method to automatically select noisy sentences and avoid manual intervention.
\end{itemize}

\section{ Acknowledgments}
	This work is supported by National Key R\&D Plan (No.2016QY03D0602), NSFC (No. 61772076 and 61751201), NSFB (No. Z181100008918002), Major Project of Zhijiang Lab (No. 2019DH0ZX01), and Open fund of BDAlGGCNEL and CETC Big Data Research Institute Co., Ltd (No. w-2018018).

\bibliographystyle{aaai}
\bibliography{aaai20}

\end{document}